\documentclass[conference]{IEEEtran}
\IEEEoverridecommandlockouts
\ifCLASSINFOpdf
\else
\fi
\usepackage{graphicx}
\usepackage{dblfloatfix}
\graphicspath{{figures/}}
\usepackage[disable]{todonotes}
\usepackage{booktabs}

\usepackage{amsmath}
\ifCLASSOPTIONcompsoc
  \usepackage[caption=false,font=normalsize,labelfont=sf,textfont=sf]{subfig}
\else
  \usepackage[caption=false,font=footnotesize]{subfig}
\fi
\usepackage{siunitx}
\usepackage{lipsum}
\usepackage{xcolor}

\newcommand{\figref}[1]{Fig.~\ref{#1}}
\newcommand{\secref}[1]{Sec.~\ref{#1}}

\newcommand{\ie}{i.\,e.\ }
\newcommand{\eg}{e.\,g.,\ }
\newcommand{\wrt}{w.\,r.\,t.\ }
\newcommand{\etal}{et\,al.\ }
\newcommand{\refhist}{\textit{RefHist}}

\newcommand{\commentOut}[1]{}

\begin{document}

%
\title{Histogram-based Deep Learning for Automotive Radar}

\author{\IEEEauthorblockN[1]{Kilian Rambach}
\IEEEauthorblockA[1]{Bosch Center for\\ Artificial Intelligence}
\and
\IEEEauthorblockN{Maxim Tatarchenko}
}%

\author{\IEEEauthorblockN{Maxim Tatarchenko*, Kilian Rambach*
} \\
\IEEEauthorblockA{
Bosch Center for Artificial Intelligence, Renningen, Germany\\
Email: maxim.tatarchenko@de.bosch.com, kilian.rambach@de.bosch.com
\thanks{*All authors have contributed equally.}
}}


%


\maketitle

\begin{abstract}
There are various automotive applications that rely on correctly interpreting point cloud data recorded with radar sensors.
We present a deep learning approach for histogram-based processing of such point clouds.
Compared to existing methods, the design of our approach is extremely simple: it boils down to computing a point cloud histogram and passing it through a multi-layer perceptron.
Our approach matches and surpasses state-of-the-art approaches on the task of automotive radar object type classification.
It is also robust to noise that often corrupts radar measurements, and can deal with missing features of single radar reflections.
Finally, the design of our approach makes it more interpretable than existing methods, allowing insightful analysis of its decisions.
\end{abstract}


%
\IEEEpeerreviewmaketitle

\section{Introduction}
Radar sensors are used in many different automotive applications including Automatic Emergency Breaking (AEB), Automatic Cruise Control (ACC), and partially automated driving.
Oftentimes these applications internally rely on the processing of radar point clouds extracted by the digital signal processing (DSP) pipelines (computing radar spectra, detecting radar reflections, estimating the direction of arrivals).
Designing such radar point cloud processing methods using machine learning is a challenging task due to the inherent data characteristics. Firstly, the number of points in each measurement cycle varies. Secondly, individual features might be missing for certain points.
Very noisy measurements or systematic errors in the
measurements can lead to a low quality of certain features,
\eg the elevation angle, and some features are not computed at all in the DSP pipeline, as a reliable computation is not possible.
In the case of low quality features, it is potentially beneficial to neglect them in decision making.

Conventionally, deep learning methods for radar point cloud processing are built around the PointNet\cite{pointnet} architecture design.
In a nutshell, it first processes per-point features independently, followed by globally aggregating the information via max-pooling to produce a point cloud embedding vector.
While such a design is applicable in many situations, it has one noticeable disadvantage: it cannot efficiently deal with missing features of individual points.

\begin{figure}
  \centering
  \includegraphics[width=\linewidth]{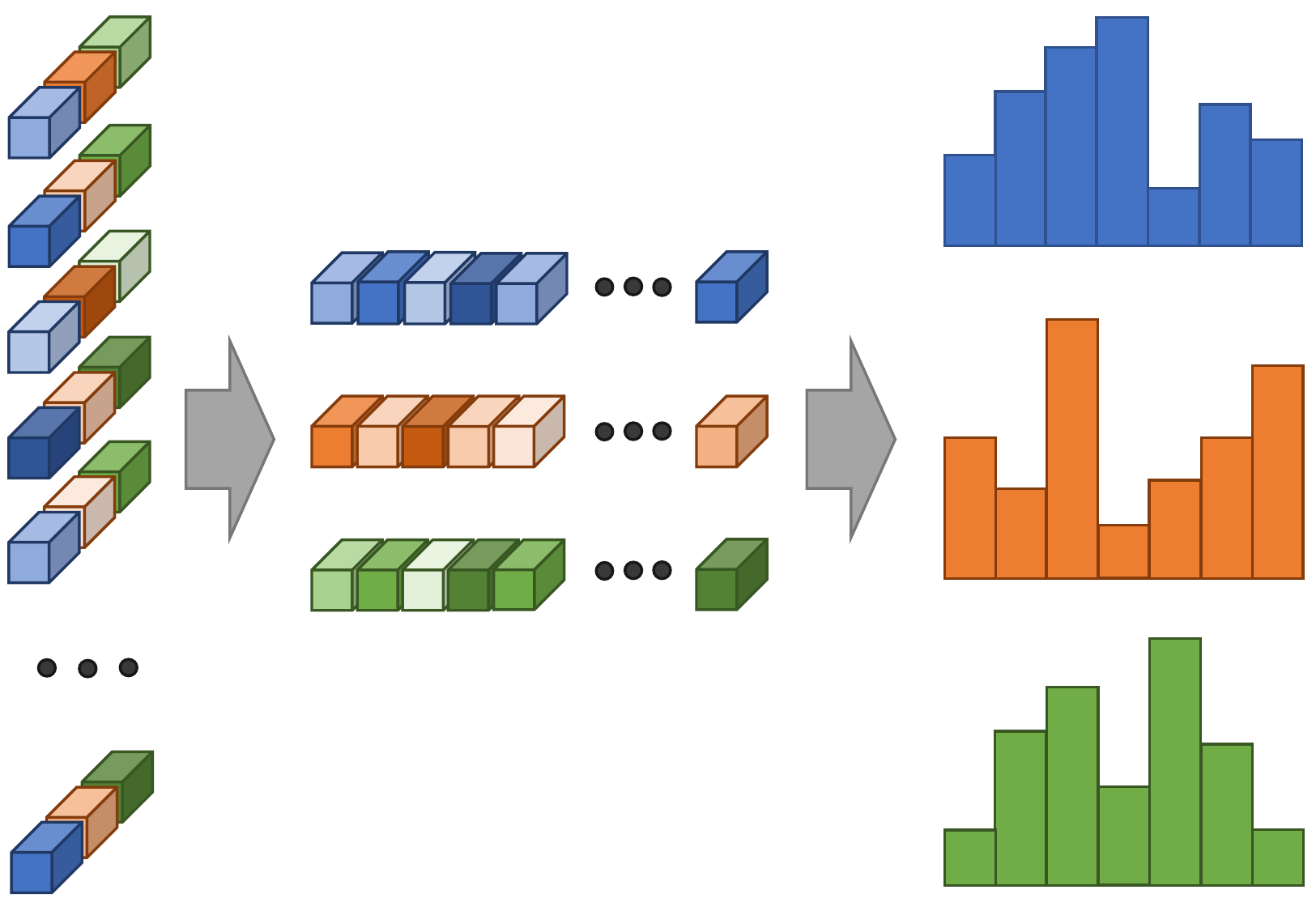}
  \caption{Our approach is based on pre-computing feature-wise histograms for the input radar point clouds. One row on the left side of the figure corresponds to the features of one radar point. Different colors in the figure correspond to different feature types (e.g. range, RCS etc.), different color intensities encode different values.}
  \label{fig:teaser}
  \vspace{-1.5em}
\end{figure}

\begin{figure*}[ht!]
  \centering
  \includegraphics[width=\linewidth]{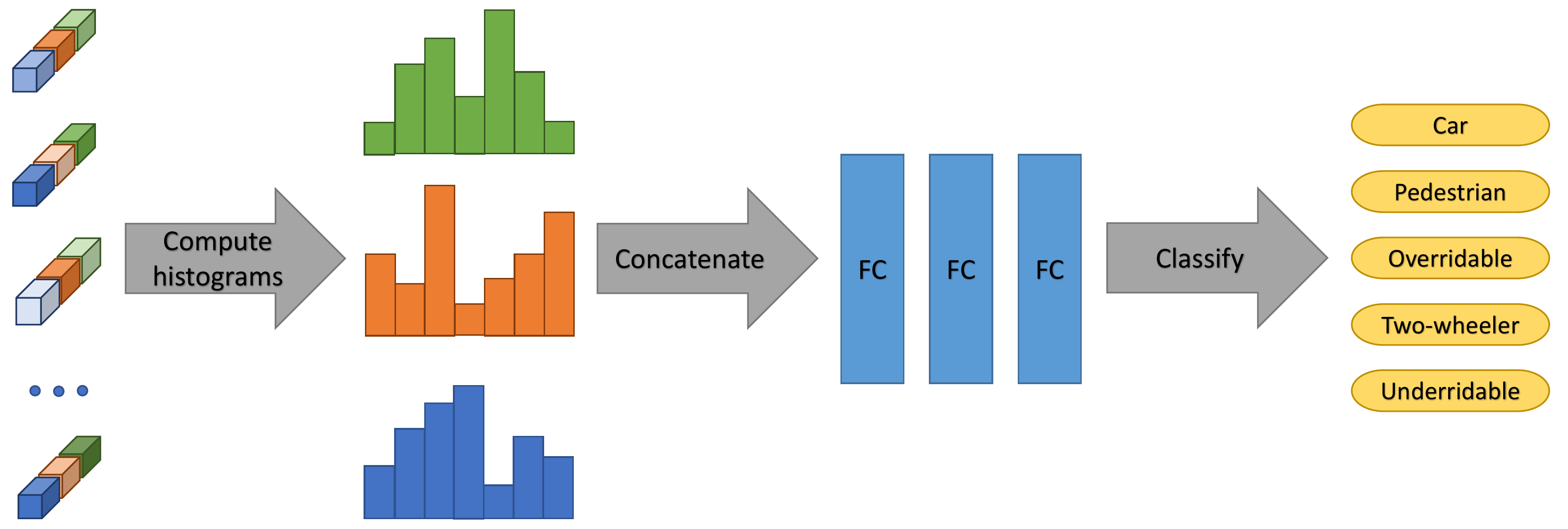}
  \caption{We aggregate the feature values for different radar data attributes (represented as different colors in the figure) by computing feature histograms. Those are used as input to a neural network with 3 fully-connected (FC) layers, which outputs the final class predictions.}
  \label{fig:pipeline_scheme}
\end{figure*}

We present an alternative deep learning approach for radar point cloud processing.
The design of our method is simple and lean: for every radar feature we compute a histogram over all points in the point cloud, see \figref{fig:teaser}.
The resulting histogram vectors are then passed through a simple multi-layer perceptron (MLP), \ie a sequence of fully-connected layers, to produce a point cloud embedding, see \figref{fig:pipeline_scheme}.
We showcase our approach on the task of radar object type classification, where it outperforms state-of-the-art methods.

Due to its simplicity, our method can be quickly implemented in any framework or on any hardware.
We study the effect of measurement noise on the final prediction accuracy and show that our method can effectively handle corrupted measurements.
This is beneficial especially for features like the azimuth angle, which become noisy for measurements with small signal-to-noise ratio (SNR).
Our approach can also naturally deal with missing features of individual points, as these simply do not contribute to the corresponding feature histograms.
These properties make our method more robust than PointNet-based architectures.
Moreover, the support of per-point removal of individual features enables an efficient prediction analysis strategy.
By discarding individual features of single points or of point groups during evaluation, and re-running the downstream predictor on those modified samples, we can assess the importance of those features for the final task.
We empirically demonstrate all these benefits on the task of radar object type classification.
Note that we get these advantages without sacrificing performance: our method achieves state-of-the-art classification accuracy in an automotive scenario.

In summary, our proposed radar point cloud processing method enjoys the following properties.

\begin{itemize}
    \item Simple and lean design, which can be quickly implemented on any hardware
    \item State-of-the-art performance on the task of radar object type classification
    \item Natural handling of missing features for individual points or point groups
    \item Robustness to additive noise
    \item Interpretability allowing to analyze and understand the predictions made by the algorithm
\end{itemize}


\section{Related Work}
In recent years different deep learning methods have been applied to automotive radar data to solve tasks like object type classification and object detection.
The methods can be categorized according to their input data type: radar reflections (point cloud data), radar spectra or a combination of both.

Radar-reflection-based methods are often inspired by the PointNet architecture family \cite{pointnet, pointNetPlusPlus}.
Ulrich \etal \cite{ulrich2020deepreflecs} use radar reflections as input to a neural network to solve an object type classification task.
Schumann \etal \cite{schumann2018SemanticSegmentationRadar} tackle semantic segmentation of radar point clouds.
Lombacher \etal \cite{ensembleLombacher2017, staticObjectClassificationDL-lombacher2016} convert radar point clouds to occupancy grids and process those with a convolutional neural network (CNN) to classify objects.
Object detection based on radar point clouds is investigated by Tilly \etal  \cite{tilly2020-object-detection-tracking-radar-point-clouds-daimler}. Ulrich \etal \cite{radarObjectDetectionHybrid-ulrich2022} combine point cloud processing with grid-based methods to further improve object detection performance.

Other approaches work directly on radar spectra.
Multiple successive range-Doppler spectra are processed by a CNN and a Long-Short-Term-Memory (LSTM) neural network \cite{convLSTM-Radar-microDoppler2019} to solve a two-class classification problem.
Patel \etal \cite{deepLearningClassificationRadarPatel2019} classify stationary objects by processing range-azimuth spectra with a CNN.
Several works investigate radar object detection based on radar spectra.
Zhang \etal \cite{raddet-object-detection-radar-sepctra-zhang2021} use range-azimuth-Doppler spectra for object detection.
Major \etal \cite{Major_2019_ICCV} optimize the computation time by collapsing either the range, azimuth, or Doppler dimensions.
Yu \etal \cite{rodnet-merging-enhancement-fusion} feed the real and imaginary parts of a sequence of range-Azimuth spectra into a neural network to detect and classify cars, pedestrians, and cyclists.
Their approach is further improved both in performance and computation time by Ju \etal in \cite{baiduObjectDetection2021}. Decourt \etal \cite{darod2022} work on range-Doppler spectra to solve an object detection task.
Rebut \etal \cite{rawHighDefRadarMultiTask} also use range-Doppler spectra, but in their approach the angular representation is learned as part of the object detection network.

Sometimes radar reflections and radar spectra are combined \cite{3dRadarcube-Cnn-classification2020, Cozma2021DeepHybrid} to achieve better object type classification performance in automotive scenarios.

Our method is most related to that of Ulrich \etal \cite{ulrich2020deepreflecs}.
It solves the same task on the same data type but offers several advantages including simpler design, better robustness and increased interpretability.

\section{Method}
\begin{figure}
  \centering
  \includegraphics[width=0.47\linewidth]{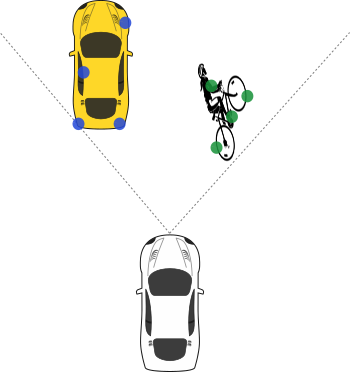}
  \caption{Measured objects and their associated radar reflections (blue points correspond to the point cloud of the car, green points correspond to the cyclist.)}
  \label{fig:topViewReflectionAssociation}
  \vspace{-1.5em}
\end{figure}

Our method receives a set of radar reflections that are associated to a single object in a single measurement cycle.
We further refer to such a reflection set as a point cloud.
Two such point clouds are illustrated in \figref{fig:topViewReflectionAssociation} with blue and green points.
Each reflection in the point cloud comes with a set of associated feature values, including for example the radial distance, azimuth angle, elevation angle, Doppler velocity, and radar cross-section (RCS).
We describe the pre-processing pipeline used to extract those feature values in more detail in \secref{sec:dataset}.
Note that fundamentally our method is not restricted to radar point clouds but can be applied to point sets coming from other sources, \eg a LiDAR sensor.

\subsection{Core approach}
Our method expects an input point cloud with N points, each being a tuple of size M, where M is the number of associated features.
For every feature, we split its effective value range (determined by the normalization step, see below) into K bins and count the number of points that fall within each bin.
This yields M histogram vectors each of length K, which we then flatten and pass through an MLP to produce the point cloud encoding.
The resulting encoding becomes an input for the downstream application.
An overview of our pipeline is shown in Fig. \ref{fig:pipeline_scheme}.
In this work, we use classification as our main application, however, the same process could be used for other tasks.
As the proposed method converts features of the radar reflections into histograms, we further refer to it as \refhist.

\begin{figure*}[ht!]
  \centering
  \includegraphics[width=\linewidth]{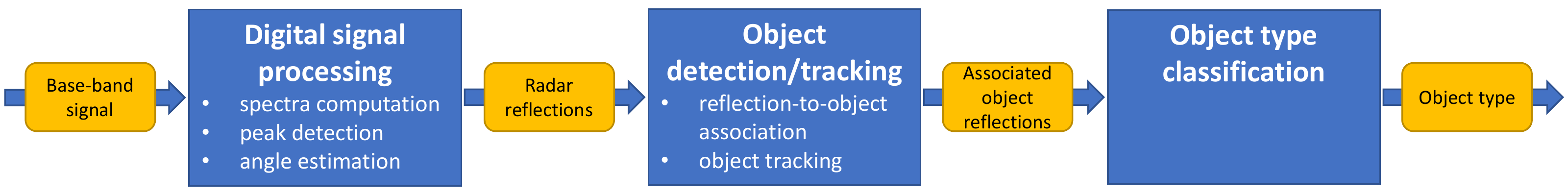}
  \caption{In the full signal processing pipeline the raw base-band signals are processed to extract per-object point clouds, which are then fed into the object type classifier.}
  \label{fig:processingPipeline}
  \vspace{-1.5em}
\end{figure*}

\subsection{Robustness}
\label{sec:method_robustness}

Since we compute the histograms separately for every feature, there is no requirement for every point to have a complete list of associated features.
This makes the method robust to situations where certain features are not available for individual points.
In reality, this can happen for example if the radar signal processing cannot compute certain features in very complex scenarios due to the limited computational resources.
Furthermore, very noisy measurements or systematic measurement errors sometimes make a reliable feature computation impossible, \eg for the elevation angles.
Approaches that directly process point cloud data rather than their aggregated statistics have a harder time dealing with such situations.
Ulrich \etal \cite{ulrich2020deepreflecs} would need to replace the missing feature values with some default values (\eg zeros) in order to not discard the "incomplete" points altogether.
This is particularly problematic when such manipulations happen only during evaluation, introducing deviations from the typical training data patterns.
Such a scenario is, however, very practically relevant: if the recording conditions for the evaluation data are not precisely the same as for the training data, we still expect the method to perform reasonably and demonstrate a certain amount of generalization.

Due to the discretization introduced by the histogram computation, our approach is also robust to additive noise.
This is especially important for features like the azimuth and elevation angle, which can become very noisy for small signal to noise ratio (SNR), \eg for objects with a large distance to the radar sensor.
We present the corresponding experiments demonstrating these properties in Sec. \ref{sec:robustness}.

\subsection{Feature normalization}

An important pre-processing step of the histogram computation is the feature normalization, i.e. which part of the feature value range is discretized into the bins.
There are multiple viable strategies one can use.

\begin{enumerate}
    \item Discretize the entire range between the minimum and the maximum feature value in the dataset. In this case, all available measurements will contribute to the histogram computation.
    \item Manually define the range boundaries, clip (i.e. assign to the boundary values) features that are outside of this defined range. This strategy is more robust to outliers. It is most sensible when there is some prior information available about the application, e.g. that the elevation angle measured by the radar cannot possibly exceed 70 degrees.
    \item Determine the effective range from the basic statistics of the data, for example $[\mu-2\sigma, \mu+2\sigma]$, where $\mu$ and $\sigma$ are the mean and the standard deviation of the feature values. The values outside of this effective range can be clipped. This is another robust strategy which can be used to remove outliers in a more data-driven fashion.
\end{enumerate}

In the end, the choice of the normalization method is determined by the data and the domain-specific knowledge about the application.
We use the data-driven normalization 3) in our experiments.

\subsection{Prediction analysis}

Our histogram-based point cloud processing supports an easy way to analyze which features of which points are the most important ones for the final prediction.
By removing individual features of individual points and re-running the predictor on such modified samples, we can assess the importance of those features for the final decision.
A similar kind of analysis can also be applied on a larger scale by removing values of a certain feature type for a pre-defined percentage of points in the entire dataset.
We demonstrate the usefulness of such analysis in Sec. \ref{sec:prediction_analysis}.
For reasons already described in Sec. \ref{sec:method_robustness}, applying the same kind of analysis in PointNet-based methods \cite{ulrich2020deepreflecs} is more problematic.
\section{Experiments}
We consider the radar object type classification problem similar to \cite{Cozma2021DeepHybrid, ulrich2020deepreflecs}.
Given the radar reflections corresponding to a single object in a single measurement cycle (point cloud for short), the task is to classify the object correctly.
We also experiment with combining the point clouds with the spectral radar data.
In the following section we describe the dataset and the data extraction in more detail.

\subsection{Dataset}
\label{sec:dataset}

A car with a front-mounted automotive radar sensor operating at 77~GHz was used to collect the measurements.
We use the recordings from both test-track scenarios (with different moving and stationary objects) and real-world drives in cities, on rural roads and on highways.
An example frame from a recording is shown in \figref{fig:sample_analysis}.
A radar signal processing pipeline is used to detect and track objects, which are then annotated with the following category labels: \textit{car}, \textit{pedestrian}, \textit{two-wheeler}, \textit{overridable} (\eg a coke-can or a manhole cover), and \textit{underridable} (\eg a bridge or a street light).
An overview of the used pipeline is depicted in  \figref{fig:processingPipeline}.
Digital radar signal processing (DSP) is used to convert the sampled base-band signal into radar spectra, detect the radar reflections using a constant false alarm rate (CFAR) detector and to estimate the angles.
Those reflections are used to detect new objects or track objects already detected in former cycles.
Therefore, each object contains several radar reflections associated to it, see \figref{fig:topViewReflectionAssociation}. Here we use a simple approach, where all radar reflections which are located inside a rectangular box around the object are associated to it. The features of the radar reflections we use are the radial distance, ego-motion-compensated radial Doppler velocity (subtracting the radial projection of the ego velocity), RCS, and the Cartesian coordinates $x,y,z$. The Cartesian coordinates are computed \wrt the center of the tracked object, similar to \cite{ulrich2020deepreflecs}. We use Cartesian coordinates in the object coordinate system instead of the spherical (range, azimuth, elevation) representation, since the object shape in Cartesian coordinates is independent of the distance between the object and the radar sensor. This facilitates the learning process of the neural network.

We split the recorded dataset into a training, validation, and test set. During the measurements, each tracked object is measured for a certain amount of time in consecutive cycles. Our splitting strategy ensures that all data samples originating from one object track are exclusively in one set, in order to avoid overfitting.
The total number of data samples is approximately 189k, where one data sample corresponds to a single point cloud.
There are 97K cars, 27K pedestrians, 31K overridables, 11K two-wheelers, and 23K underridables.
The training, validation, and test set contain $70\%, 20\%$, and $10\%$ of the data, respectively.

In additional experiments, we use point cloud data combined with spectral radar data around the object reflections as an input to the neural network.
For a detailed description of the radar signal processing used in this case, refer to \cite{Cozma2021DeepHybrid}. We use the same architecture as proposed in \cite{Cozma2021DeepHybrid}, but substitute the point cloud branch by our proposed method.

\commentOut{
\begin{itemize}
    \item car with automotive radar sensor 77 Ghz
    \item real measurements on a test track and in real world in a city, rural roads and on highways.
    \item objects are labeled and consist of the classes "car", "pedestrian", "two-wheeler", "overridable" like a coke-can or manhole cover, and "underridable", e.g. a bridge or street lights. Those are relevant for automotive radar applications.
    \item radar processing, radar reflections with attributes like azimuth angle, elevation angle, distance, Doppler velocity, RCS are computed
    \item objects are detected and tracked, and radar reflections are associated to it
    \item task: given the associated point of an object, classify the object
    \item features are the Cartesian coordinates x, y, z, radial distance, ego motion compensated Doppler velocity and RCS of the radar reflection. x,y,z are computed \wrt the center ob the tracked object. see deepreflecs paper for more details
    \item splitting of dataset: objects are measured for a certain amount of time several time. Splitting strategy ensures that all data samples originating from one object track are exclusively in one set to avoid overfitting. 
    \item splitting percentages
    \item number samples
    \item we also run experiments to combine the point cloud data with spectral radar data. This is done by extracting parts of the spectra around the detected radar reflections and using them as additional input to the neural network. For a detailed description refer to \cite{Cozma2021DeepHybrid}.
    
\end{itemize}
} 

\subsection{Training details}
Every per-feature histogram consists of 20 bins.
We input those into a 3-layer MLP to compute the final class predictions, each of the inner network layers having 16 output neurons.
We experiment with different network sizes in Sec. \ref{sec:hyperparameter_search}.
We train all our networks for 1000 epochs using the Adam optimizer \cite{kingma2015adam} with the learning rate set to $10^{-5}$.
The batch size is 64.
The model is trained using the standard cross-entropy classification loss.
To account for the imbalanced number of samples per class in the dataset, we additionally scale the per-sample loss values with the following class-dependent weights

\begin{equation}
    w_i = \frac{N_\text{samples}}{N_\text{classes}\cdot N_i},
\end{equation}

where $N_\text{samples}$ is the total number of samples in the dataset, $N_\text{classes}$ is the total number of classes and $N_i$ is the number of samples for class $i$.

\subsection{Main results}

\begin{table*}
  \centering
  \begin{tabular}{@{}lccccccc@{}}
    \toprule
    Method & Car & Pedestrian & Overridable & Two-wheeler & Underridable & Balanced accuracy & \# parameters\\
    \midrule
    DeepReflecs & 0.90 & 0.83 & 0.84 & 0.80 & 0.95 & 0.87 & 2989\\
    RefHist (ours) & \textbf{0.92} & \textbf{0.87} & \textbf{0.87} & \textbf{0.87} & \textbf{0.96} & \textbf{0.90} & \textbf{2293}\\
    \midrule
    DeepReflecs + Spectra & 0.97 & \textbf{0.89} & \textbf{0.90} & 0.94 & 0.95 & 0.93 & 32429\\
    RefHist (ours) + Spectra & \textbf{0.98} & \textbf{0.89} & \textbf{0.90} & \textbf{0.95} & \textbf{0.96} & \textbf{0.94} & \textbf{32069}\\
    \bottomrule
  \end{tabular}
  \vspace{0.6em}
  \caption{Quantitative comparison of our method with DeepReflecs \cite{ulrich2020deepreflecs}.}
  \label{tbl:quantitative_results}
  \vspace{-3em}
\end{table*}

We compare our method to DeepReflecs \cite{ulrich2020deepreflecs}, the current state-of-the-art approach for radar object type classification, see Tbl. \ref{tbl:quantitative_results}.
We slightly modified their original architecture by adding an extra fully-connected layer before the classifier, which in all our experiments proved to be better than the vanilla DeepReflecs.
Our approach outperforms DeepReflecs both in terms of overall balanced accuracy and on individual classes.
Note that this result is achieved without sacrificing the efficiency.
In fact, our network is smaller than DeepReflecs, see the '\# parameters' column in Tbl. \ref{tbl:quantitative_results}.
The same observations hold when combining the point clouds and spectra in a single model: an approach using our histogram branch outperforms the one using DeepReflecs.
This demonstrates that our method can be used successfully as part of larger processing pipelines.

Note that the core of our method is not the network architecture itself but rather the way we pre-process and interpret the input data.
Instead of preserving per-point correspondences between individual features as done in DeepReflecs, we replace the raw point clouds with high-level feature statistics.
The fact that such a seemingly severe transformation does not lead to any performance drop compared to DeepReflecs indicates that point-wise correspondences between features may not be crucial for the task or not actually be exploited effectively by the DeepReflecs architecture, even despite being provided in the input data.
Working with high-level statistics, however, has certain advantages in terms of robustness and interpretability, which we study in the further experiments.

\subsection{Robustness}
\label{sec:robustness}
In this section we analyze the robustness of our method to additive noise and to partially missing features.
Having a robust algorithm is important, in particular when the test-time recording conditions differ from those under which the training data was collected.

Firstly, we randomly sample zero-mean normally distributed noise with different standard deviations $\sigma$, and add it to all features in the point cloud during evaluation.
We consider two amounts of noise: $\sigma=0.0125$ and $\sigma=0.025$, the latter corresponding to half of the histogram bin width.
Tbl. \ref{tbl:robustness} summarizes the quantitative performance of both our method and DeepReflecs.
Note that the methods were only trained on "clean" data, i.e. we only add noise at test time.

\begin{table}[ht]
  \centering
  \begin{tabular}{@{}lccc@{}}
    \toprule
    Method & W/o noise & $\sigma=0.0125$ & $\sigma=0.025$ \\
    \midrule
    DeepReflecs & 0.866 & 0.819 & 0.773 \\
    RefHist (ours) & \textbf{0.900} & \textbf{0.896} & \textbf{0.883} \\
    \bottomrule
  \end{tabular}
  \vspace{0.6em}
  \caption{Robustness to noise.}
  \vspace{-1.5em}
  \label{tbl:robustness}
\end{table}

Our approach proves to be significantly more robust in all regimes.
For larger noise, its overall balanced accuracy drops by 3\% as compared to the 10\% drop of DeepReflecs.

Secondly, we evaluate the robustness of our approach to missing features of individual points.
We simulate this by randomly removing a certain percentage of values for a single feature from the entire test dataset. For DeepReflecs, we replace those values by $0$, since they cannot be simply removed, as discussed in \secref{sec:method_robustness}.
The results of this experiment for different percentages of removed feature values are summarized in Tbl. \ref{tbl:missing_features}.

\begin{table}[ht]
  \centering
  \begin{tabular}{@{}lccc@{}}
    \toprule
    Method & Original & $5\%$ & $90\%$ \\
    \midrule
    DeepReflecs & 0.866 & 0.431 & 0.228 \\
    RefHist (ours) & \textbf{0.900} & \textbf{0.899} & \textbf{0.881} \\
    \bottomrule
  \end{tabular}
  \vspace{0.6em}
  \caption{Robustness to removed feature values.}
  \vspace{-1.5em}
  \label{tbl:missing_features}
\end{table}

Note how already 5\% of removed feature values effectively destroy DeepReflecs's performance, while our method's overall accuracy stays undisturbed.
With as much as 90\% of the feature values removed, our method still delivers excellent accuracy, while DeepReflecs's performance drops to chance level.
It is worth mentioning that the resulting performance decrease naturally depends on the importance of the specific feature for the final prediction.
Here, the feature we removed is the Cartesian component $y$, i.e. the lateral displacement of the reflection \wrt the object center, which happens to be relatively unimportant.
Still, removing 5\% of such an unimportant cue turns out to be sufficient to bring down DeepReflecs.
We further discuss the matter in the next section and demonstrate that removing other features might have a more significant effect on the method's performance.

\subsection{Interpretability}
\label{sec:prediction_analysis}
\begin{figure}
  \centering
  \includegraphics[width=\linewidth]{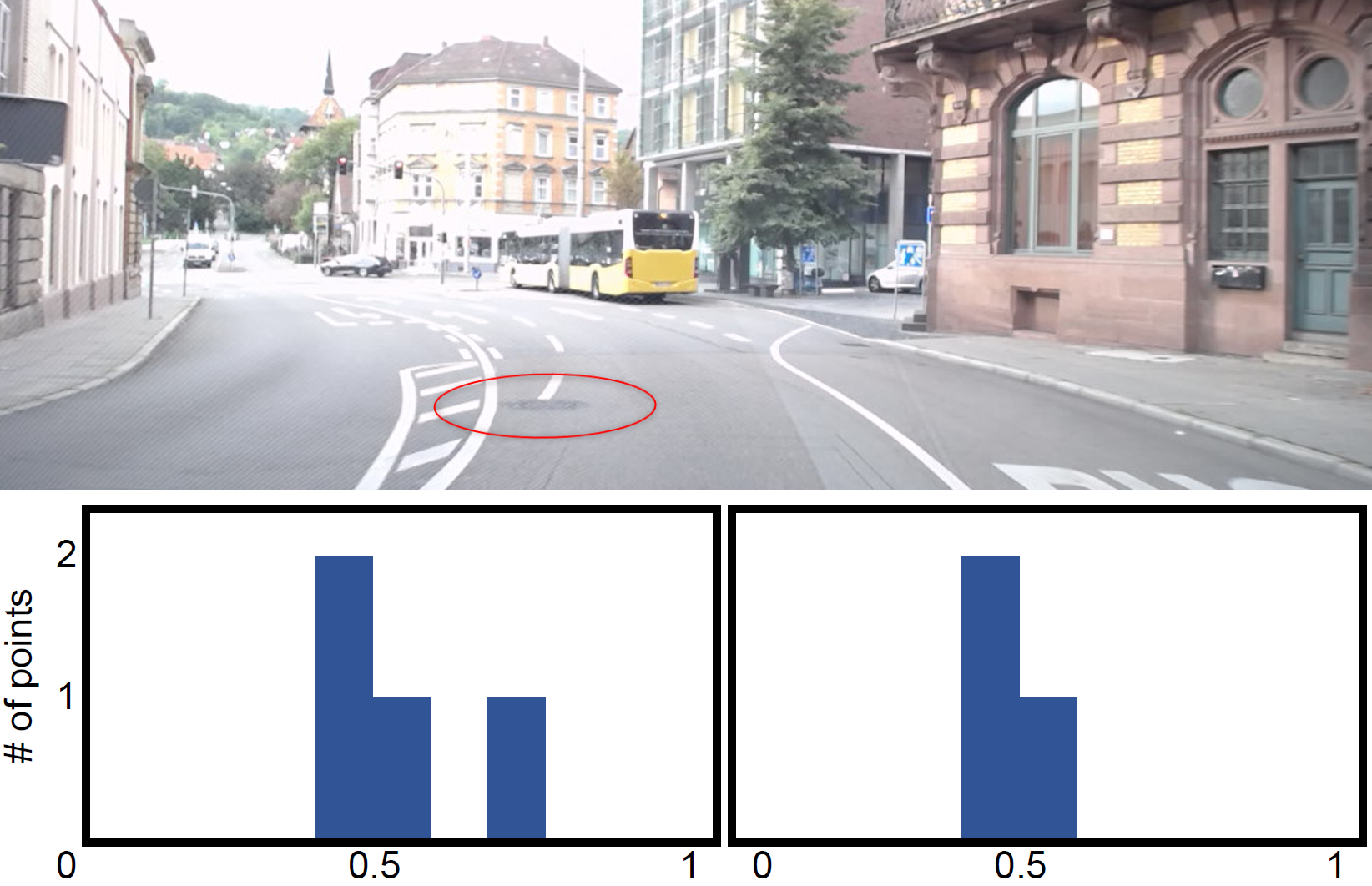}
  \caption{Removing the largest elevation value from the wrongly classified left histogram (corresponding to the overridable manhole cover in the top image) results in the histogram on the right, which is classified correctly.}
  \label{fig:sample_analysis}
  \vspace{-1.5em}
\end{figure}

Our method's robustness to the missing features of single points allows us to derive a simple strategy for prediction analysis.
On the sample level, we can identify individual feature values that contribute to the final decision by removing them and re-running the resulting histogram through the predictor.
An example of such an analysis is shown in Fig. \ref{fig:sample_analysis}.

The bottom-left histogram corresponds to the point cloud recording of the overridable object, a manhole cover, shown on the top.
The sample was misclassified as underridable by the algorithm.
However, removing the largest elevation feature value, which corresponds to the bottom-right histogram, makes the algorithm predict the correct class label.
Such per-sample analysis can be valuable for situational understanding of the algorithm's behavior.
As discussed in Sec. \ref{sec:method_robustness}, applying the same analysis on DeepReflecs is not trivial.

\begin{figure}
  \centering
  \includegraphics[width=\linewidth]{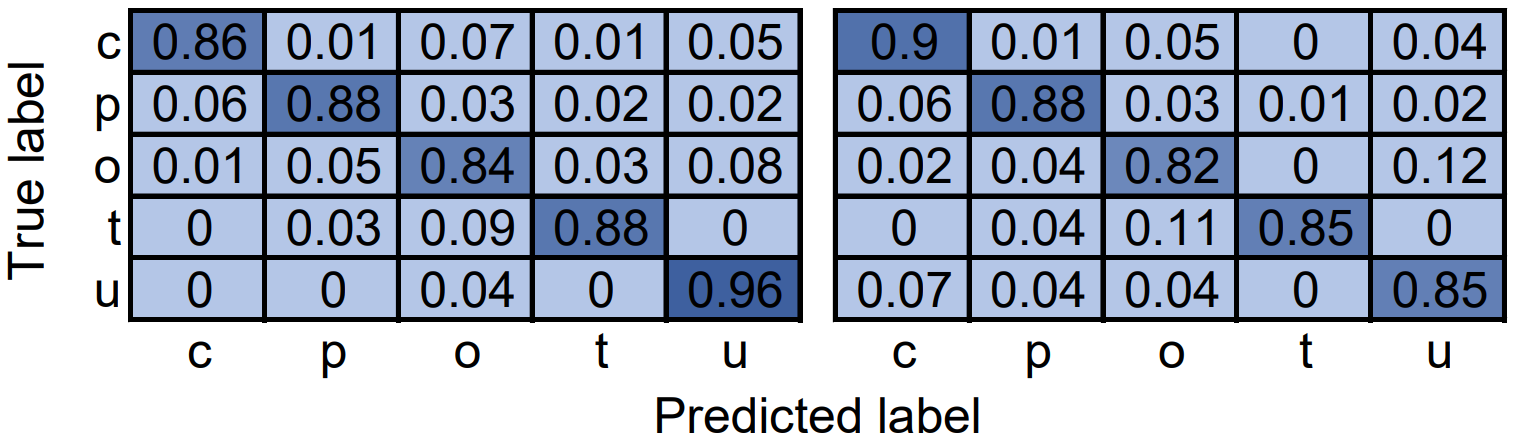}
  \caption{Confusion matrices for point clouds with removed features. Left: 90\% of $y$ feature removed, right: 90\% of elevation feature removed. Class abbreviations: c (car), p (pedestrian), o (overridable), t (two-wheeler), u (underridable). Elevation is more important than $y$ for the final classification performance, especially for the overridable class.}
  \label{fig:removed_features_confusion_matrices}
\end{figure}

A similar kind of analysis can be applied on a larger scale.
Fig. \ref{fig:removed_features_confusion_matrices} shows confusion matrices for two evaluation regimes: one with 90\% of the lateral displacement feature values removed (Cartesian component $y$), and the other one with 90\% of the elevation feature values removed (Cartesian component $z$).
One can see that generally elevation has a much higher effect on the overall performance than the lateral displacement.
Moreover, it shows that the missing elevation feature leads to higher confusion of underridables with car, pedestrian, or overridables.
In general, overridables have a much larger elevation value than the other classes.
Removing this information leads to increased confusion.
Such insights can provide valuable information for planning the data collection campaigns or better understanding the internal functioning of the algorithm.
They can also give hints on which aspects of the sensor and its DSP are important for the classification task - an information that can be exploited in future sensor development.

\subsection{Hyperparameter search}
\label{sec:hyperparameter_search}

In the previous experiments we used a medium-sized model that delivered sufficient performance for our problem.
Generally, there is a trade-off between the model size and the prediction accuracy.
To better understand this trade-off, we explored the space of models in a basic hyperparameter search.

We used a three-layer MLP with fixed input/output dimensions as our backbone, which effectively leaves two parameters that affect the model size: the output dimensions of the first two layers.
In Tbl. \ref{tbl:hyperparameter_search} one can see how changing those affects the final prediction accuracy and the overall number of parameters.
The performance across different models is very stable and it monotonously increases with the increased model size.
In the end, the model selection is usually informed by each application's hardware constraints.

\begin{table}
  \centering
  \begin{tabular}{@{}lccc@{}}
    \toprule
    Layer 1 & Layer 2 & Performance & \# parameters\\
    \midrule
    4 & 4 & 0.86 & 529\\
    4 & 8 & 0.87 & 569\\
    4 & 16 & 0.87 & 649\\
    8 & 4 & 0.88 & 1029\\
    8 & 8 & 0.89 & 1085\\
    8 & 16 & 0.89 & 1197\\
    8 & 32 & 0.89 & 1421\\
    16 & 4 & 0.89 & 2029\\
    16 & 8 & 0.89 & 2117\\
    16 & 16 & 0.90 & 2293\\
    16 & 32 & 0.90 & 2645\\
    32 & 8 & 0.90 & 4181\\
    32 & 16 & 0.90 & 4485\\
    32 & 32 & 0.91 & 5093\\
    \bottomrule
  \end{tabular}
  \vspace{0.6em}
  \caption{Hyperparameter search.}
  \label{tbl:hyperparameter_search}
  \vspace{-1.5em}
\end{table}

\commentOut{
\begin{table}
  \centering
  \begin{tabular}{@{}lccc@{}}
    \toprule
    Layer 1 & Layer 2 & Performance & \# parameters\\
    \midrule
    4 & 4 & 0.88 & 1493\\
    4 & 8 & 0.89 & 2793\\
    4 & 16 & 0.90 & 5393\\
    8 & 4 & 0.90 & 1537\\
    8 & 8 & 0.90 & 2853\\
    8 & 16 & 0.91 & 5485\\
    16 & 4 & 0.89 & 1625\\
    16 & 8 & 0.91 & 2973\\
    16 & 16 & 0.91 & 5669\\
    \bottomrule
  \end{tabular}
  \vspace{0.6em}
  \caption{Hyperparameter search.}
  \label{tbl:hyperparameter_search}
  \vspace{-1.5em}
\end{table}
}
\section{Conclusions}

We presented an approach for histogram-based radar point cloud processing suitable for automotive applications.
Despite its simplicity, our approach outperforms state-of-the-art methods on radar object type classification while demonstrating better robustness and interpretability than the competing approaches.

In this paper we focused solely on extracting high-level point cloud statistics represented as histograms.
However, per-point feature correspondences, when used effectively, might contain useful cues missing from our representation.
Investigating better ways to exploit such per-point correspondences and combining those with statistical representations is an important direction of future research.

\section*{Acknowledgments}

The authors thank David Stöckel, Felix Kleber, Jonathan Kirchner, Christian Wegend, Christian Weiss, Gor Hakobyan, Johannes Fink, Arthur Hipke, and William Beluch for fruitful discussions and support in data and manuscript preparation.




\bibliographystyle{IEEEtran}
\bibliography{literature}

\begin{thebibliography}{10}
\providecommand{\url}[1]{#1}
\csname url@samestyle\endcsname
\providecommand{\newblock}{\relax}
\providecommand{\bibinfo}[2]{#2}
\providecommand{\BIBentrySTDinterwordspacing}{\spaceskip=0pt\relax}
\providecommand{\BIBentryALTinterwordstretchfactor}{4}
\providecommand{\BIBentryALTinterwordspacing}{\spaceskip=\fontdimen2\font plus
\BIBentryALTinterwordstretchfactor\fontdimen3\font minus
  \fontdimen4\font\relax}
\providecommand{\BIBforeignlanguage}[2]{{%
\expandafter\ifx\csname l@#1\endcsname\relax
\typeout{** WARNING: IEEEtran.bst: No hyphenation pattern has been}%
\typeout{** loaded for the language `#1'. Using the pattern for}%
\typeout{** the default language instead.}%
\else
\language=\csname l@#1\endcsname
\fi
#2}}
\providecommand{\BIBdecl}{\relax}
\BIBdecl

\bibitem{pointnet}
R.~Q. {Charles}, H.~{Su}, M.~{Kaichun}, and L.~J. {Guibas}, ``Pointnet: Deep
  learning on point sets for 3d classification and segmentation,'' in
  \emph{CVPR}, 2017.

\bibitem{pointNetPlusPlus}
C.~R. Qi, L.~Yi, H.~Su, and L.~J. Guibas, ``Pointnet++: Deep hierarchical
  feature learning on point sets in a metric space,'' in \emph{NeurIPS}, 2017.

\bibitem{ulrich2020deepreflecs}
M.~Ulrich, C.~Gläser, and F.~Timm, ``Deepreflecs: Deep learning for automotive
  object classification with radar reflections,'' in \emph{RadarConf}, 2021.

\bibitem{schumann2018SemanticSegmentationRadar}
O.~Schumann, M.~Hahn, J.~Dickmann, and C.~Wohler, ``Semantic segmentation on
  radar point clouds,'' in \emph{FUSION}, 2018.

\bibitem{ensembleLombacher2017}
J.~{Lombacher}, M.~{Hahn}, J.~{Dickmann}, and C.~{Wöhler}, ``Object
  classification in radar using ensemble methods,'' in \emph{ICMIM}, 2017.

\bibitem{staticObjectClassificationDL-lombacher2016}
J.~Lombacher, M.~Hahn, J.~Dickmann, and C.~Woehler, ``Potential of radar for
  static object classification using deep learning methods,'' in \emph{ICMIM},
  2016.

\bibitem{tilly2020-object-detection-tracking-radar-point-clouds-daimler}
J.~F. Tilly, S.~Haag, O.~Schumann, F.~Weishaupt, B.~Duraisamy, J.~Dickmann, and
  M.~Fritzsche, ``Detection and tracking on automotive radar data with deep
  learning,'' in \emph{FUSION}, 2020.

\bibitem{radarObjectDetectionHybrid-ulrich2022}
M.~Ulrich, S.~Braun, D.~Köhler, D.~Niederlöhner, F.~Faion, C.~Gläser, and
  H.~Blume, ``Improved orientation estimation and detection with hybrid object
  detection networks for automotive radar,'' in \emph{ITSC}, 2022.

\bibitem{convLSTM-Radar-microDoppler2019}
H.-U.-R. Khalid, S.~Pollin, M.~Rykunov, A.~Bourdoux, and H.~Sahli,
  ``Convolutional long short-term memory networks for doppler-radar based
  target classification,'' in \emph{RadarConf}, 2019.

\bibitem{deepLearningClassificationRadarPatel2019}
K.~Patel, K.~Rambach, T.~Visentin, D.~Rusev, M.~Pfeiffer, and B.~Yang, ``Deep
  learning-based object classification on automotive radar spectra,'' in
  \emph{RadarConf}, 2019.

\bibitem{raddet-object-detection-radar-sepctra-zhang2021}
A.~Zhang, F.~E. Nowruzi, and R.~Laganiere, ``Raddet: Range-azimuth-doppler
  based radar object detection for dynamic road users,'' in \emph{CRV}, 2021.

\bibitem{Major_2019_ICCV}
B.~Major, D.~Fontijne, A.~Ansari, R.~Teja~Sukhavasi, R.~Gowaikar, M.~Hamilton,
  S.~Lee, S.~Grzechnik, and S.~Subramanian, ``Vehicle detection with automotive
  radar using deep learning on range-azimuth-doppler tensors,'' in \emph{ICCV
  Workshops}, 2019.

\bibitem{rodnet-merging-enhancement-fusion}
J.~Yu, X.~Hao, X.~Gao, Q.~Sun, Y.~Liu, P.~Chang, Z.~Zhang, F.~Gao, and
  F.~Shuang, ``Radar object detection using data merging, enhancement and
  fusion,'' in \emph{ICMR}, 2021.

\bibitem{baiduObjectDetection2021}
B.~Ju, W.~Yang, J.~Jia, X.~Ye, Q.~Chen, X.~Tan, H.~Sun, Y.~Shi, and E.~Ding,
  ``Danet: Dimension apart network for radar object detection,'' in
  \emph{ICMR}, 2021.

\bibitem{darod2022}
C.~Decourt, R.~VanRullen, D.~Salle, and T.~Oberlin, ``Darod: A deep automotive
  radar object detector on range-doppler maps,'' in \emph{IV}, 2022.

\bibitem{rawHighDefRadarMultiTask}
J.~Rebut, A.~Ouaknine, W.~Malik, and P.~Pérez, ``Raw high-definition radar for
  multi-task learning,'' in \emph{CVPR}, 2022.

\bibitem{3dRadarcube-Cnn-classification2020}
A.~{Palffy}, J.~{Dong}, J.~F.~P. {Kooij}, and D.~M. {Gavrila}, ``Cnn based road
  user detection using the 3d radar cube,'' \emph{RA-L}, 2020.

\bibitem{Cozma2021DeepHybrid}
A.-E. Cozma, L.~Morgan, M.~Stolz, D.~Stoeckel, and K.~Rambach, ``Deephybrid:
  Deep learning on automotive radar spectra and reflections for object
  classification,'' in \emph{ITSC}, 2021.

\bibitem{kingma2015adam}
D.~P. Kingma and J.~Ba, ``Adam: A method for stochastic optimization,'' in
  \emph{ICLR}, 2015.

\end{thebibliography}
%


%

\end{document}